\documentclass[10pt,twocolumn,letterpaper]{article}

\usepackage{iccv}
\usepackage{times}
\usepackage{epsfig}
\usepackage{graphicx}
\usepackage{amsmath}
\usepackage{amssymb}
\usepackage[table]{xcolor} 
\usepackage{bm}
\usepackage{bbm}

\usepackage{booktabs}
\usepackage{multirow}
\usepackage{arydshln}

\definecolor{darkergreen}{RGB}{21, 152, 56}

\newcommand{\tabincell}[2]{\begin{tabular}{@{}#1@{}}#2\end{tabular}}  

\usepackage[pagebackref=true,breaklinks=true,letterpaper=true,colorlinks,bookmarks=false]{hyperref}

\iccvfinalcopy 

\ificcvfinal\fi

\begin{document}

\title{SoftCLIP: Softer Cross-modal Alignment Makes CLIP Stronger}

\author{%
  Yuting Gao$^{1}$\thanks{The first three authors contributed equally. This work was done when J. Liu was an intern at Tencent. }
  \quad Jinfeng Liu$^{1,2, *}$ \quad Zihan Xu$^{1, *}$ \\ Tong Wu$^{1}$  \quad Enwei Zhang$^{1}$ \quad Wei Liu$^{2}$  \quad Jie Yang$^{2}$ \quad Ke Li$^{1}$ \quad Xing Sun$^{1}$ 
  \\[0.25cm]
\small{$^1$Tencent Youtu Lab
 \quad $^2$Shanghai Jiaotong University 
}
\\
{
    \tt\small 
    \{yutinggao,ianxxu\}@tencent.com,ljf19991226@sjtu.edu.cn
}
}

\maketitle
\ificcvfinal\thispagestyle{empty}\fi

\begin{abstract}

   During the preceding biennium, vision-language pre-training has achieved noteworthy success on several downstream tasks. Nevertheless, acquiring high-quality image-text pairs, where the pairs are entirely exclusive of each other, remains a challenging task, and noise exists in the commonly used datasets. To address this issue, we propose SoftCLIP, a novel approach that relaxes the strict one-to-one constraint and achieves a soft cross-modal alignment by introducing a softened target, which is generated from the fine-grained intra-modal self-similarity. The intra-modal guidance is indicative to enable two pairs have some local similarities and model many-to-many relationships between the two modalities. Besides, since the positive still dominates in the softened target distribution, we disentangle the negatives in the distribution to further boost the relation alignment with the negatives in the cross-modal learning. Extensive experiments demonstrate the effectiveness of SoftCLIP. In particular, on ImageNet zero-shot classification task, using CC3M/CC12M as pre-training dataset, SoftCLIP brings a top-1 accuracy improvement of 6.8\%/7.2\% over the CLIP baseline.
\end{abstract}

\section{Introduction}

Since OpenAI proposed Contrastive Language-Image Pre-training (CLIP)~\cite{radford2021learning}, large-scale vision-language pre-training (VLP) has achieved rapid development. Many approaches~\cite{li2021supervision, yao2021filip, gao2022pyramidclip, li2021align} have been proposed and achieved remarkable success on several downstream tasks. 

\begin{figure}[t]
    \centering
    \includegraphics[width=8.2cm]{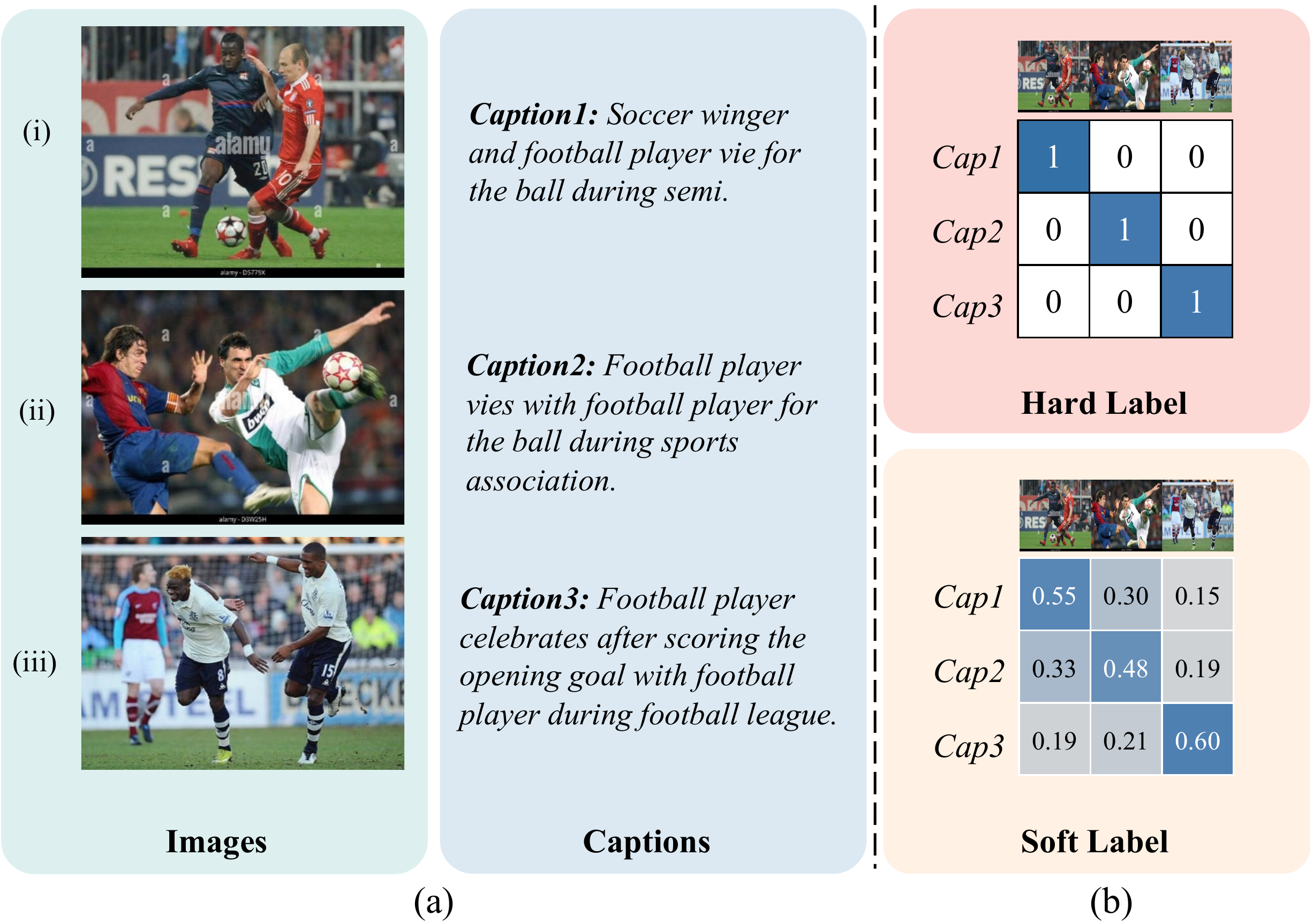}
    \caption{(a) Three image-text pairs randomly sampled from CC3M dataset have some local similarities, suggesting the ubiquitous many-to-many relationships. (b) Using fine-grained intra-modal self-similarity as the softened target can allow for the existence of some similarities among unpaired image and text.}
    \label{figure1}
\end{figure}

Among these methods, the alignment of the visual and linguistic modalities is a critical component, often requiring the use of image-text contrastive learning. This learning process aims to bring paired image and text samples closer while simultaneously pushing unpaired samples away, necessitating the complete mutual exclusivity between any two unpaired samples.
However, acquiring high-quality image-text pairs is a challenging task, owing to the fact that the majority of image-text pairs are obtained through web crawling over the Internet, which frequently results in significant noise. As evidenced in Figure~\ref{figure1}(a), there are some local similarities between the three pairs, the caption of (i) can also be used to describe the image (ii) and (iii), indicating many-to-many relationships instead of perfect one-to-one correspondences, which is also pointed out in CLIP-PSD~\cite{andonian2022robust}. Therefore, it is too harsh and unreasonable to completely push away the image (i) and the text (ii)/(iii). Recent work PyramidCLIP~\cite{gao2022pyramidclip} also noticed this problem and proposed to use label smoothing~\cite{szegedy2016rethinking} to mitigate this problem. However, assigning equal weight to all the negative samples is improper and ignores the information pertaining to their relationships. The neglect of the potential distinctions among negative samples results in the underutilization of valuable information and an incomplete understanding of the underlying data structure.

In this paper, we propose SoftCLIP, a novel approach that relaxes the strict one-to-one contrastive constraint and leverages the intra-modal discriminative information to guide the interaction between visual and linguistic modalities. Specifically, we employ fine-grained intra-modal self-similarities as the softened targets for soft cross-modal alignments. Figure~\ref{figure1}(b) illustrates how our softened targets allow for the existence of some similarities between the image (i) and the text (ii)/(iii). By incorporating the softened targets, SoftCLIP overcomes the limitations of traditional contrastive methods and captures the nuanced information between visual and linguistic modalities, leading to a significant improvement in cross-modal learning. Furthermore, treating different negative samples with different weights helps the model to capture the authentic distribution of data more effectively. However, the contribution of negatives in the softened target distribution can still be overwhelmed by the dominant positive one. To address this problem, we take further step to disentangle the negatives in the distribution. Specifically, we sift out the negative logits regardless of the positive logit in both prediction and target distributions with renormalization, and then bring the new two distributions closer, which boosts the relation alignment with negatives and brings further improvement.

Extensive experiments on several downstream tasks demonstrate the effectiveness of the proposed SoftCLIP. Specifically, using CC3M~\cite{changpinyo2021conceptual}/CC12M~\cite{sharma2018conceptual} as pre-training dataset and ResNet50~\cite{He_2016_CVPR}-Transformer~\cite{vaswani2017attention} as the image-text encoder, SoftCLIP achieved 24.2\%/43.2\% top-1 accuracy on zero-shot ImageNet~\cite{deng2009imagenet} classification task, which is 6.8\%/7.2\% higher than its baseline CLIP.

Our main contributions are summarized as follows:

\begin{itemize}

\item We propose to employ fine-grained intra-modal self-similarities as softened targets for cross-modal learning, thereby alleviating the problem of non-strict mutual exclusion between any two pairs. 

\item We boost the relation alignment with negatives by disentangling the negatives in the distribution to alleviate them being overwhelmed by the positive one. 

\item We also use symmetric KL-Divergence to replace the conventional cross-entropy when incorporating the softened targets. Extensive experiments demonstrate the effectiveness of SoftCLIP, which can steadily bring significant improvements under various scales of pre-training data and various model architectures. 

\end{itemize}

\section{Related Work}

\subsection{Vision Language Pre-training}

Vision-language pretraining (VLP) strives to achieve a unified representation of two modalities, namely vision and language, through the utilization of large-scale image-text pairs.  Existing VLP models can be broadly categorized based on their architectures into three types, \textit{i.e.}, dual-stream models for alignment, single-stream models for fusion, or their combination.

As a paradigmatic dual-stream model, CLIP~\cite{radford2021learning} has exhibited remarkable performance on zero-shot recognition and several downstream tasks by leveraging contrastive learning on large-scale image-text pairs. Following this paradigm, SLIP~\cite{mu2022slip} and DeCLIP~\cite{li2021supervision} further combine self-supervision to improve data utilization efficiency. PyramidCLIP~\cite{gao2022pyramidclip} and FILIP~\cite{yao2021filip} introduce finer-grained and more interactions between two modalities, seeking for more accurate cross-modal alignment. CyCLIP~\cite{goel2022cyclip} points out the importance of geometric consistency in the learned representation space between two modalities, and proposes geometrically consistency constraints. Different from dual-stream ones, single-stream models, such as Visual-BERT~\cite{li2019visualbert} and OSCAR~\cite{li2020oscar}, fuse the image and text features with a unified model to achieve a deeper interaction. ALBEF~\cite{li2021align} and CoCa~\cite{yu2022coca} absorb the essence of the two kinds of structures, and find a more flexible way to learn visual and linguistic representations. In this paper, we adopt the dual-stream architecture and depart from the commonly used hard one-hot labels. Instead, we utilize fine-grained intra-modal self-similarities as softened targets to provide more informative guidance, which leads to improved cross-modal interactions.

\subsection{Softened Target}
Softened target aims to alleviate the strict constraints imposed by one-hot label and avoid the model's overconfidence towards wrong predictions, which has demonstrated its effectiveness across various tasks. For example, label smoothing~\cite{szegedy2016rethinking}, a commonly used strategy in classification task, assigns some small positive values to the ground-truth of all negative samples. Moreover, in the field of knowledge distillation~\cite{hinton2015distilling}, the logits predicted by the teacher model will be used as softened targets to guide the learning process of student model. The softened targets,  containing the teacher's modeling of the relationship among all the samples, are more instructive than the one-hot label. Recently, PyramidCLIP~\cite{gao2022pyramidclip} has pointed out the potential limitation of the overly rigid one-hot label, and hence proposes to use label smoothing to mitigate this problem. However, it should be emphasized that the indiscriminate treatment towards all negative samples is unreasonable and necessitates further attention. CLIP-PSD~\cite{andonian2022robust} also utilizes softened targets obtained from a teacher model to reduce the adverse effects of noisy image-text pairs. Its core concept is progressive self-distillation where the student network acts as its own teacher and the model dynamically evolves into its own teacher as training progresses. From this perspective, SoftCLIP is also working under the self-distillation framework, however, the softened targets do not stem from the images and texts, but from the pre-extracted ROI (region-of-interest) features of objects and corresponding tags.

\begin{figure*}[ht]
    \centering
    \includegraphics[width = 1.0\textwidth]{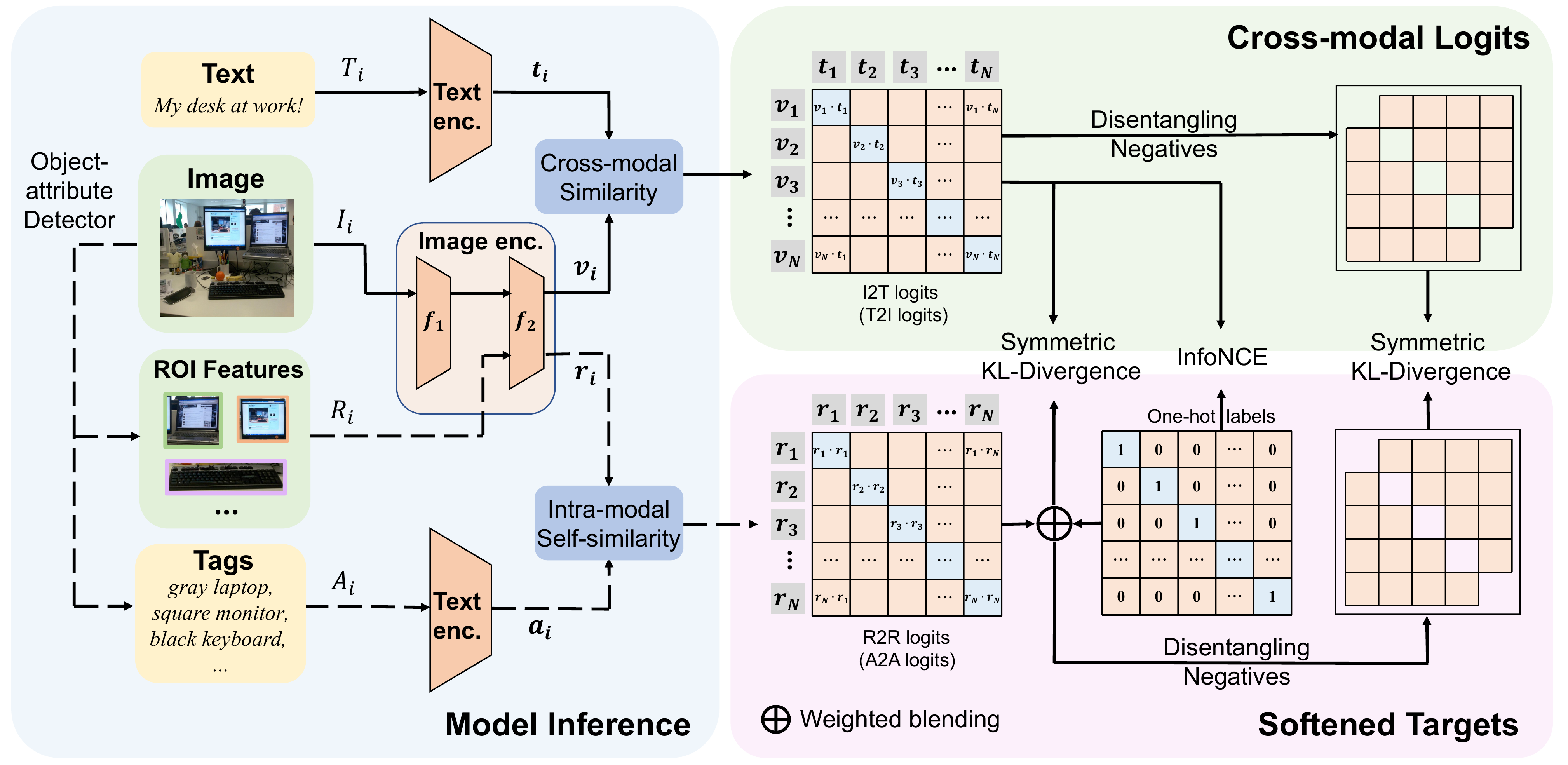}
    \caption{
    The overall framework of SoftCLIP. For each image-text pair, the image is fed into a pre-trained object-attribute detector to extract ROI features and their corresponding tags, which are used to compute the intra-modal self-similarities to guide the cross-modal interactions. Besides, we disentangle negatives in each distribution to construct another soft loss term and boost the relation alignment with negatives. And the conventional cross-entropy is replaced by symmetric KL-Divergence when incorporating the softened targets.}
    \label{framework}
\end{figure*}

\section{Methodology}

In this section, we first present some CLIP preliminaries, and then introduce the details of our proposed SoftCLIP. The overall framework can be seen in Figure \ref{framework}.

\subsection{CLIP Preliminaries and Label Smoothing}
Consider a batch of $N$ image-text pairs $\{(I_i, T_i)\}_{i=1}^N$, CLIP employs a dual-stream encoder to obtain the semantic representation of each pair. Specifically, for the $i_{th}$ pair, the image data $I_i$ is input into an image encoder to get the visual representation $\bm{v}_i$, and the text data $T_i$ is input into a text encoder to get the linguistic representation $\bm{t}_i$, generating L2-normalized embedding pairs $\{(\bm{v}_i, \bm{t}_i)\}_{i=1}^N$. CLIP uses InfoNCE~\cite{oord2018representation} to conduct cross-modal alignment, which pulled the paired image and text embeddings together while pushing unpaired apart. For the $i_{th}$ pair, the normalized image-to-text similarity vector $\bm{p}_{i}(I, T)=\{p_{ij}(I, T)\}_{j=1}^N$ and the text-to-image counterpart $\bm{p}_{i}(T, I)=\{p_{ij}(T, I)\}_{j=1}^N$ can be calculated through:
\vspace{-5pt}
\begin{align}
p_{ij}(I, T) =\frac{\mathrm{exp}(\mathrm{sim}(\bm{v_i},\bm{t_j})/\tau)}{\sum_{j=1}^N \mathrm{exp}(\mathrm{sim}(\bm{v_i},\bm{t_j})/\tau)},
\end{align}
\vspace{-10pt}
\begin{align}
p_{ij}(T, I) =\frac{\mathrm{exp}(\mathrm{sim}(\bm{t_i},\bm{v_j})/\tau)}{\sum_{j=1}^N \mathrm{exp}(\mathrm{sim}(\bm{t_i},\bm{v_j})/\tau)},
\end{align}
where $\tau$ is a learnable temperature parameter initialized with $0.07$ and the function $\mathrm{sim}( \cdot )$ conducts dot product to measure the similarity scores. In CLIP paradigm, the corresponding one-hot label vectors are used as the targets to calculate InfoNCE loss. The one-hot label of the $i_{th}$ pair is denoted as $\bm{y}_{i}=\{y_{ij}\}_{j=1}^N$, with $y_{ii}$ equal to $\mathrm{1}$ and all other elements equal to $\mathrm{0}$. Therefore the vision-to-language loss and the language-to-vision loss can be obtained by:
\vspace{-5pt}
\begin{align}
\mathcal{L}_{v2l} =\frac{1}{N} \sum_{i=1}^N H(\bm{y}_{i}, \bm{p}_{i}(I, T)),
\end{align}
\vspace{-12pt}
\begin{align}
\mathcal{L}_{l2v} =\frac{1}{N} \sum_{i=1}^N H(\bm{y}_{i}, \bm{p}_{i}(T, I)),
\end{align}
where $H(\cdot, \cdot)$ denotes the cross-entropy operation.
And the final CLIP loss can be denoted as $\mathcal{L}_{\rm CLIP} = (\mathcal{L}_{v2l}+\mathcal{L}_{l2v})/2$.

As we have discussed, CLIP neglects some local similarities between unpaired images and texts within a batch, while PyramidCLIP roughly uses label smoothing to soften the hard one-hot targets to alleviate this issue. Specifically, the original one-hot label vector $\bm{y}_{i}$ is softened to $\widetilde {\bm{y}}_{i}$, which is formulated as:
\begin{align}
{\widetilde {\bm y} }_{i} = (1-\alpha) \bm{y}_{i} + \frac{\alpha}{N-1} (\mathbbm{1}-\bm{y}_{i}),
\end{align}
where $\alpha$ is the smoothing hyper-parameter set to $0.2$, and $\mathbbm{1}$ denotes the all-ones vector.

\subsection{Soft Alignment under Intra-modal Guidance}
The label smoothing strategy transfers a small portion of the confidence from the positive sample and amortizes it to the negatives, allowing for weak and fixed similarity with negatives. This strategy works in PyramidCLIP, however, the improvement it brings is limited since it merely models naive many-to-many relationships between images and the corresponding texts. To improve this, we try to find clues from the relation within a single modality. Specifically, we attempt to use the intra-modal self-similarity as the softened target to guide the CLIP model. An accurate intra-modal self-similarity can provide a superb supervision to repair a sample with more semantically similar correspondences in another modality. Moreover, it inherently contains the implicit expression of many-to-many relationships, with rich and instructive knowledge.

Intuitively, we may choose the original images and texts to calculate the intra-modal self-similarity, \textit{i.e.}, image-to-image similarity for the visual modality and text-to-text similarity for the textual modality. However, this approach encounters some problems and does not perform well in practice, which is revealed in the experimental part. PyramidCLIP pre-extracts the ROI features of detected salient objects for each image, with tag description for each object, to introduce cross-level relation alignment, which can bring significant gains. The ROI features and corresponding tags of objects, extracted by a pre-trained object-attribute detector, contain the prior category and attribute information of objects from the task of object detection. This encourages us to exploit the priors, \textit{i.e.}, we can alternatively use the ROI features and tags to calculate the intra-modal self-similarity.

Formally, for the image-text pair $(I_i, T_i)$, we can pre-extract the corresponding ROI-tag (ROI features and tags) pair $(R_i, A_i)$ from the image $I_i$, constructing ROI-tag pairs $\{(R_i, A_i)\}_{i=1}^N$ within a batch. Note that 
the tags are concatenated and separated by commas to form a sentence. Each pair is feed into the dual-stream model following PyramidCLIP. As shown in Figure \ref{framework}, $R_i$ is processed by the rear part of the image encoder and $A_i$ is processed by the text encoder, deriving the corresponding L2-normalized representation vector pairs $\{(\bm{r}_i, \bm{a}_i)\}_{i=1}^N$. And the linear embedding layers for transforming vector dimension are omitted here. For the $i_{th}$ pair, the normalized intra-modal self-similarity vectors of $R_i$ and $A_i$, denoted as $\bm{p}_{i}(R, R)=\{p_{ij}(R, R)\}_{j=1}^N$ and $\bm{p}_{i}(A, A)=\{p_{ij}(A, A)\}_{j=1}^N$ respectively, can be obtained by:
\vspace{-5pt}
\begin{align}
p_{ij}(R, R) =\frac{\mathrm{exp}(\mathrm{sim}(\bm{r_i},\bm{r_j})/\tau)}{\sum_{j=1}^N \mathrm{exp}(\mathrm{sim}(\bm{r_i},\bm{r_j})/\tau)},
\end{align}
\vspace{-15pt}
\begin{align}
p_{ij}(A, A) =\frac{\mathrm{exp}(\mathrm{sim}(\bm{a_i},\bm{a_j})/\tau)}{\sum_{j=1}^N \mathrm{exp}(\mathrm{sim}(\bm{a_i},\bm{a_j})/\tau)}.
\end{align}

Next, the ROI self-similarity and tag self-similarity are utilized as the soft labels to supervise the image-to-text and text-to-image correspondences respectively. In practice, we use the weighted average of the hard labels and the soft labels as the final softened targets to ensure the training stability and better generalization, which is formulated as:
\vspace{-4pt}
\begin{align}
\label{eqmix1}
{\widetilde {\bm p} }_{i}(R, R) = (1-\beta) \bm{y}_{i} + \beta \bm{p}_{i}(R, R),
\end{align}
\vspace{-15pt}
\begin{align}
\label{eqmix2}
{\widetilde {\bm p} }_{i}(A, A) = (1-\beta) \bm{y}_{i} + \beta \bm{p}_{i}(A, A),
\end{align}
where $\bm{y}_{i}$ denotes the hard one-hot label and $\beta$ is a mixing coefficient set to 0.3. Since the softened targets are also variable distributions, the cross-entropy in CLIP should be replaced by the KL-Divergence as follows:
\vspace{-5pt}
\begin{align}
\mathcal{L}_{\mathrm{soft}\text{-}v2l} =\frac{1}{N} \sum_{i=1}^N \mathrm{KL}({\widetilde {\bm p} }_{i}(R, R)\ ||\ \bm{p}_{i}(I, T)),
\end{align}
\vspace{-15pt}
\begin{align}
\mathcal{L}_{\mathrm{soft}\text{-}l2v} =\frac{1}{N} \sum_{i=1}^N \mathrm{KL}({\widetilde {\bm p} }_{i}(A, A)\ ||\ \bm{p}_{i}(T, I)).
\end{align}
Then we can get the average soft loss under the guidance of ROIs and tags, denoted as $\mathcal{L}_{\mathrm{soft}}=(\mathcal{L}_{\mathrm{soft}\text{-}v2l}+\mathcal{L}_{\mathrm{soft}\text{-}l2v})/2$.

\subsection{Boosting Relation Alignment with Negatives}
The introducing of intra-modal self-similarity does relax the strict one-to-one constraint and guide the model to learn many-to-many correspondences between the visual and linguistic modalities. However, the confidence of the positive sample still dominates compared to the negatives despite of the softened target distribution. This may lead to numerous negatives submerged by the dominant positive ones in the cross-modal relation alignment. And the problem will be more serious when meeting faulty positives, which means the paired images and texts in the web-harvested dataset are actually irrelevant. To mitigate this issue, we disentangle negatives in the distribution to boost the relation alignment with negatives in SoftCLIP.

\begin{figure}[h]
    \centering
    \includegraphics[width=8cm]{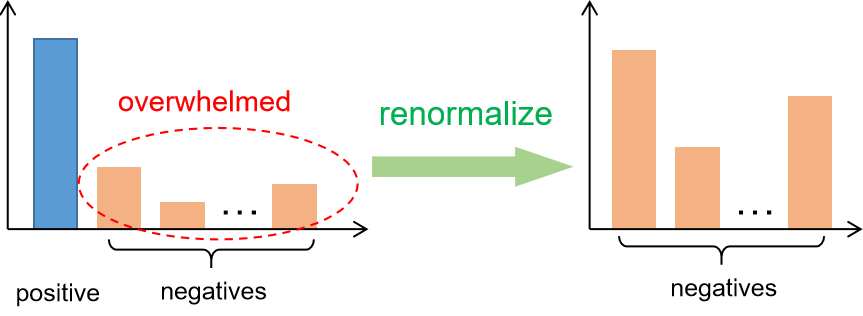}
    \caption{Disentangling the negatives in the distribution.}
    \label{nckd}
\end{figure}

Specifically, we discard the positive logits in the probability distribution and only concentrate on the knowledge among negative logits with renormalization, as shown in Figure \ref{nckd}. For any distribution vector ${\bm p}_{i}=\{p_{ij}\}_{j=1}^N \in \mathbb R^{ 1\times N}$, we use ${\bm p}_{i}^{*}=[p_{i1}^*, ...,p_{i(i-1)}^*,p_{i(i+1)}^*, ..., p_{iN}^*] \in \mathbb R^{ 1\times (N-1)}$ to denote its corresponding neg-disentangled distribution, with the elements calculated through:
\begin{align}
p_{ij}^* = \frac{p_{ij}}{\sum_{k=1,k\neq i}^N p_{ik}},
\end{align}
where $j$ is taken from $[1,...,i-1,i+1,...,N]$. The disentangling of negatives is applied identically to the distributions ${\widetilde {\bm p} }_{i}(R, R)$, ${\widetilde {\bm p} }_{i}(A, A)$, ${\bm p}_i (I,T)$ and ${\bm p}_i (T,I)$, generating ${\widetilde {\bm p} }_{i}^*(R, R)$, ${\widetilde {\bm p} }_{i}^*(A, A)$, ${\bm p}_i^* (I,T)$ and ${\bm p}_i^* (T,I)$ correspondingly. Then we can derive the relation-enhanced formulation of $\mathcal{L}_{\mathrm{soft}\text{-}v2l}$ and $\mathcal{L}_{\mathrm{soft}\text{-}l2v}$ as:
\vspace{-5pt}
\begin{align}
\mathcal{L}_{\mathrm{soft}\text{-}v2l}^{\mathrm{re}} =\frac{1}{N} \sum_{i=1}^N \mathrm{KL}({\widetilde {\bm p} }_{i}^*(R, R)\ ||\ \bm{p}_{i}^*(I, T)),
\end{align}
\vspace{-15pt}
\begin{align}
\mathcal{L}_{\mathrm{soft}\text{-}l2v}^{\mathrm{re}} =\frac{1}{N} \sum_{i=1}^N \mathrm{KL}({\widetilde {\bm p} }_{i}^*(A, A)\ ||\ \bm{p}_{i}^*(T, I)).
\end{align}
Hence, the relation-enhanced soft loss can be written as $\mathcal{L}_{\mathrm{soft}}^{\mathrm{re}}=(\mathcal{L}_{\mathrm{soft}\text{-}v2l}^{\mathrm{re}}+\mathcal{L}_{\mathrm{soft}\text{-}l2v}^{\mathrm{re}})/2$.

\subsection{Training Objective}
It is well known that the KL-Divergence is essentially asymmetric, whereas the JS-Divergence is an alternative with symmetric form. However, we have observed that the JS-Divergence makes the training stage unstable. Therefore, we directly symmetrize the KL-Divergence by adding a reversed term with the two input distributions exchanged, which has been proved to be effective in the experiments. For instance, the symmetric form of $D=\mathrm{KL}({\bm p}\ ||\ {\bm q})$ can be written as:
\begin{align}
\widetilde{D} =\frac{1}{2} (\mathrm{KL}({\bm p}\ ||\ {\bm q})+\mathrm{KL}({\bm q}\ ||\ {\bm p})).
\end{align}
Following this, we can symmetrize $\mathcal{L}_{\mathrm{soft}}$ and $\mathcal{L}_{\mathrm{soft}}^{\mathrm{re}}$, obtaining $\widetilde{\mathcal{L}}_{\mathrm{soft}}$ and $\widetilde{\mathcal{L}}_{\mathrm{soft}}^{\mathrm{re}}$ respectively. And we utilize the two terms to regulate the original CLIP loss. So the overall loss function is denoted as:
\begin{align}
\mathcal{L}_{\rm SoftCLIP} = \widetilde{\mathcal{L}}_{\mathrm{soft}} + \lambda \widetilde{\mathcal{L}}_{\mathrm{soft}}^{\mathrm{re}} + \mu \mathcal{L}_{\rm CLIP},
\end{align}
where the loss weights $\lambda$ and $\mu$ are set to $1.0$ and $0.5$ in the experiments.

\section{Experiments}

\subsection{Pre-training and Evaluation Details}

\noindent\textbf{Architectures and Pre-training Datasets} SoftCLIP accommodates three distinct model architectures, with the visual encoder compatible with ResNet50, ViT-B/32~\cite{dosovitskiy2020image} and ViT-B/16~\cite{dosovitskiy2020image}, while the language encoder utilizes Transformer following CLIP~\cite{radford2021learning}. The input resolution of image encoder is 224$\times$224 and the maximum context length of text encoder is 77. And SoftCLIP is pre-trained on three datasets,  CC3M~\cite{changpinyo2021conceptual}, CC12M~\cite{sharma2018conceptual} and YFCC15M-V2~\cite{li2021supervision}.



\noindent\textbf{Object-attribute Detector} The object-attribute detector used to extract ROI features with tags is pre-trained by VinVL~\cite{zhang2021vinvl}, adopting the framework of Faster R-CNN~\cite{ren2015faster}. 
Through the detector, we take 10 objects with the highest confidence from each image to obtain the corresponding ROI features and category descriptions with attribute information. Each ROI feature is of 2052-dimension, concatenated by a 2048-dimensional appearance feature vector and 4-dimensional position vector (the coordinates of top-left and bottom-right corners of the object region).

\noindent\textbf{Implementation Details} We train our SoftCLIP using an AdamW~\cite{loshchilov2017decoupled} optimizer and the cosine learning rate scheduler with a linear warm-up. Specifically, the learning rate linearly increases from 0 to the peak value within $10\%$ of the total steps, and then decreases with a  cosine anneal strategy. The weight decay rate of AdamW is set to 0.2. To save GPU memory, automatic mixed-precision~\cite{micikevicius2018mixed} is used. The models are trained from scratch for either 8 or 32 epochs in our experiments, \textit{i.e.}, 8 epochs for ablation and 32 epochs for comparison. We use 8 V100 GPUs for experiments, when training with ResNet50 and ViT-B/32 image encoder, the batch size is set to 2048,  while with the image encoder ViT-B/16, the batch size is 1024.

\begin{table}[b]
\caption{Comparison against CLIP baseline on ImageNet. ZS denotes Zero-Shot and YFCC15M is YFCC15M-V2 dataset.}
\label{imagenetZS}
\vspace{2pt}
\setlength{\belowcaptionskip}{1pt}
\centering
\setlength{\tabcolsep}{2.5mm}{
\begin{tabular}{cccc}
\toprule
\multirow{2}{*}{\textbf{Method}} & \textbf{Pretrain} & \textbf{Image} & \textbf{ImageNet} \\  
  & \textbf{Dataset}  & \textbf{Encoder} & \textbf{ZS Top1}  \\
\midrule
CLIP$\Diamond$ & CC3M & \multirow{2}{*}{ResNet50}&  17.7 \\
\textbf{SoftCLIP} & CC3M &  &  \textbf{24.2(\textcolor{darkergreen}{+6.5})}   \\
\hdashline
CLIP$\Diamond$ & CC3M & \multirow{2}{*}{ViT-B/32}  & 11.9   \\ 
\textbf{SoftCLIP} & CC3M &  &  \textbf{13.3(\textcolor{darkergreen}{+1.4})}   \\
\hdashline
CLIP$\Diamond$ & CC3M & \multirow{2}{*}{ViT-B/16}   & 16.9   \\
\textbf{SoftCLIP} & CC3M &  & \textbf{18.9(\textcolor{darkergreen}{+2.0})}   \\
\bottomrule
CLIP$\Diamond$ & CC12M & \multirow{2}{*}{ResNet50}&  36.0 \\
\textbf{SoftCLIP} & CC12M &  &  \textbf{43.2(\textcolor{darkergreen}{+7.2})}   \\
\hdashline
CLIP$\Diamond$ & CC12M & \multirow{2}{*}{ViT-B/32}  & 31.5   \\ 
\textbf{SoftCLIP} & CC12M &  &  \textbf{34.4(\textcolor{darkergreen}{+2.9})}   \\
\hdashline
CLIP$\Diamond$ & CC12M & \multirow{2}{*}{ViT-B/16}   & 36.8  \\
\textbf{SoftCLIP} & CC12M &  & \textbf{42.1(\textcolor{darkergreen}{+5.3})}   \\
\bottomrule
CLIP$\Diamond$ & YFCC15M & \multirow{2}{*}{ResNet50}&  39.6 \\
\textbf{SoftCLIP} & YFCC15M &  &  \textbf{43.7(\textcolor{darkergreen}{+4.1})}   \\
\hdashline
CLIP$\Diamond$ & YFCC15M & \multirow{2}{*}{ViT-B/32}  & 33.1   \\ 
\textbf{SoftCLIP} & YFCC15M &  &  \textbf{35.0(\textcolor{darkergreen}{+1.9})}   \\
\hdashline
CLIP$\Diamond$ & YFCC15M & \multirow{2}{*}{ViT-B/16}  & 38.9  \\
\textbf{SoftCLIP} & YFCC15M  & & \textbf{42.4(\textcolor{darkergreen}{+3.5})}   \\
\bottomrule
\specialrule{0em}{1pt}{1pt}
\multicolumn{3}{l}{\small $\Diamond$ Our Implementation~~~~~~ }
\end{tabular}}
\end{table}

\noindent\textbf{Downstream Tasks for Evaluation} We validate the effectiveness of the proposed SoftCLIP on three downstream tasks: zero-shot image classification, zero-shot image-text retrieval and image retrieval. For zero-shot image classification, experiments are carried out on 7 datasets, such as ImageNet~\cite{deng2009imagenet}, Pets~\cite{data_pets}, Describable Textures~\cite{data_dtd}, Food-101~\cite{data_food}, Flowers-102~\cite{data_flower}, SUN397~\cite{data_sun} and Caltech-101~\cite{data_cal}. For zero-shot image-text retrieval, experiments are conducted on Flickr30K~\cite{hodosh2013framing} and MS-COCO~\cite{lin2014microsoft}. For image retrieval, two sub-tasks are included: instance retrieval task on Oxford~\cite{philbin2007object} and Paris Buildings datasets~\cite{philbin2008lost}, and copy detection task on the INRIA Copydays~\cite{douze2009evaluation} dataset.

\subsection{Zero-shot Image Classification}

To validate the effectiveness of the proposed SoftCLIP, we first conduct experiments on the widely used zero-shot ImageNet classification task. The results are presented in Table~\ref{imagenetZS}. It is clear that SoftCLIP brings significant improvement compared to the CLIP baseline with different image encoders, across varying levels of pre-training data. Notably, SoftCLIP exhibits a significant increase of 6.5\%/7.2\% in top-1 accuracy compared to CLIP when the pre-training dataset is CC3M/CC12M and the visual encoder is ResNet50.
Besides, we also provide the zero-shot classification results on the other six small datasets, which are illustrated in Table~\ref{transfer2smallcls}. Obviously, the performance of SoftCLIP significantly exceed the CLIP baseline across all the six datasets, which demonstrates the efficacy and generalization of the proposed SoftCLIP.


\begin{table}[t]
\caption{Accuracy on 6 datasets with ResNet50 and ViT-B/16 image encoder pretrained on YFCC15M-V2. PETS / DTD / F101 / FLOW / SUN / CAL are abbreviations for Pets / Describable Textures / Food-101 / Flowers-102 / SUN397 / Caltech-101 datasets. AVG represents average accuracy across all 6 datasets.}
\label{transfer2smallcls}
\setlength{\belowcaptionskip}{1pt}
\centering
\footnotesize

\setlength{\tabcolsep}{0.7mm}{
\begin{tabular}{ccccccccc}
\toprule
\textbf{Method} & \textbf{\tabincell{c}{\textbf{Image} \\ \textbf{Encoder}}} &  \textbf{PETS} &  \textbf{DTD}  &  \textbf{F101} & \textbf{FLOW} & \textbf{SUN} & \textbf{CAL} &\textbf{AVG}\\ 
\midrule
CLIP$\Diamond$  &  \multirow{2}{*}{ResNet50} & 33.3 & 22.8  & 48.0   & 54.9 & 50.0 & 65.6 & 45.8       \\
\textbf{SoftCLIP} &   & \textbf{34.9}  & \textbf{27.1} &  \textbf{50.8} & \textbf{56.3} & \textbf{55.9} & \textbf{70.4}  & \textbf{49.2} \\ 
\midrule
CLIP$\Diamond$  &  \multirow{2}{*}{ViT-B/16} & 27.2 & 21.6  & 48.3   & 53.8 & 53.4 & 71.5 & 46.0       \\
\textbf{SoftCLIP} &   & \textbf{32.5}  & \textbf{25.6} &  \textbf{53.8} & \textbf{55.6} & \textbf{56.2} & \textbf{71.8}  & \textbf{49.2} \\ 
\bottomrule
\specialrule{0em}{1pt}{1pt}
\multicolumn{6}{l}{\small $\Diamond$ Our Implementation}
\end{tabular}}
\end{table}

\subsection{Image Retrieval}
In this section, we evaluate the transferability of the SoftCLIP's visual encoder on two image retrieval tasks, {\it i.e.,} instance retrieval and copy detection. 


\noindent\textbf{Instance Retrieval} We conduct instance retrieval on two public benchmarks: Oxford and Paris Buildings datasets with revisited annotations~\cite{radenovic2018revisiting}. This task aims to retrieve the same building from the gallery as the query image. Our results are reported in terms of Mean Average Precision (mAP) on the Easy (E), Medium (M), and Hard (H) splits of these two datasets, where retrieval difficulty increases gradually from E to H. As shown in Table~\ref{insretrieval}, the proposed SoftCLIP significantly outperforms CLIP baseline over all the three data splits, confirming the robustness and transferability of SoftCLIP's visual representations.



\begin{table}[htp]
\caption{Instance retrieval results. All the models are pre-trained on YFCC15M-V2 dataset with ResNet50 as visual encoder.}
\label{insretrieval}
\setlength{\belowcaptionskip}{1pt}
\centering

\vspace{2pt}
\setlength{\tabcolsep}{2.5mm}{
\begin{tabular}{cccc}
\toprule
\textbf{Method} & \textbf{Pretrain} & \textbf{CLIP}$\Diamond$ & \textbf{SoftCLIP} \\  
\midrule
\multirow{3}{*}{\bm{$\mathcal{R}$}\textbf{Oxf}} & E & 42.0 &  \textbf{45.5(\textcolor{darkergreen}{+3.5})} \\
 & M & 28.5 & \textbf{33.3(\textcolor{darkergreen}{+4.8})}    \\
 & H & 7.0 & \textbf{11.9(\textcolor{darkergreen}{+4.9})}    \\
 \midrule
\multirow{3}{*}{{\bm{$\mathcal{R}$}\textbf{Par}}} & E & 82.5 & \textbf{85.1(\textcolor{darkergreen}{+2.6})} \\
 & M & 66.7 &  \textbf{72.6(\textcolor{darkergreen}{+5.9})}   \\
 & H & 41.0 &  \textbf{47.0(\textcolor{darkergreen}{+6.0})}   \\
\bottomrule
\specialrule{0em}{1pt}{1pt}
\multicolumn{3}{l}{\small $\Diamond$ Our Implementation~~~~~~ }
\end{tabular}}
\end{table}

\noindent\textbf{Copy Detection} We conduct the copy detection task on a ``strong" subset of the INRIA Copydays dataset, which consists of images that have undergone significant distortions. The objective of copy detection is to match the distorted, secondary edited image with its original image. As listed in Table~\ref{copydetection}, SoftCLIP still shows superiority over CLIP baseline, highlighting the generalization of SoftCLIP.

\begin{table}[t]
\caption{Results on the copy detection task. YFCC15M actually denotes YFCC15M-V2 dataset.}
\label{copydetection}
\vspace{2pt}
\setlength{\belowcaptionskip}{1pt}
\centering
\setlength{\tabcolsep}{2.5mm}{
\begin{tabular}{cccc}
\toprule
\multirow{2}{*}{\textbf{Method}} & \textbf{Pretrain} & \textbf{Image} & \multirow{2}{*}{\textbf{mAP}} \\  
  & \textbf{Dataset}  & \textbf{Encoder}  \\
\midrule
CLIP$\Diamond$ & YFCC15M & \multirow{2}{*}{ResNet50}&  74.9 \\
\textbf{SoftCLIP} & YFCC15M &  &  \textbf{77.0(\textcolor{darkergreen}{+2.1})}   \\
\bottomrule
\specialrule{0em}{1pt}{1pt}
\multicolumn{3}{l}{\small $\Diamond$ Our Implementation~~~~~~ }
\end{tabular}}
\end{table}

\subsection{Zero-shot Image-text Retrieval}
\begin{table*}[ht]
\caption{Zero-shot image-text retrieval results on Flicker30K and MS-COCO. All the models are pre-trained on YFCC15M-V2 dataset.}
\label{sotaretrieval}
\setlength{\belowcaptionskip}{2pt}
\centering
\setlength{\tabcolsep}{1.5mm}{
\begin{tabular}{cccccccccccccccc}
\toprule
\multirow{5}{*}[5pt]{\textbf{Method}}  & \multirow{5}{*}[5pt]{\tabincell{c}{\textbf{Image} \\ \textbf{Encoder}}} & \multicolumn{6}{c}{\textbf{Flickr30K(1K)}}  & \multicolumn{6}{c}{\textbf{MS-COCO(5K)}}  \\  
\cmidrule(r){3-8}\cmidrule(r){9-14}
& & \multicolumn{3}{c}{\textbf{Image-to-Text}}  & \multicolumn{3}{c}{\textbf{Text-to-Image}} & \multicolumn{3}{c}{\textbf{Image-to-Text}} & \multicolumn{3}{c}{\textbf{Text-to-Image}} & \\
\cmidrule(r){3-5}\cmidrule(r){6-8}\cmidrule(r){9-11}\cmidrule(r){12-14}
 & & R@1 & R@5 & R@10 & R@1 & R@5 & R@10 &  R@1 & R@5 & R@10 &  R@1 & R@5 & R@10  \\ 
\midrule
CLIP$\Diamond$ &\multirow{3}{*}{ResNet50}   & 54.9 & 81.6 & 90.5 & 37.1 & 65.0 & 75.0 & 29.4 & 54.8 & 66.1 & 18.9 & 40.7 & 52.5 \\ 
DECLIP$^\dagger$ &   & 58.7 & 85.0 & 92.5 & 40.7 & 68.9 & 78.4 & 31.1 & 59.0 & 69.9 & 20.6 & 43.8 & 55.4 \\
\textbf{SoftCLIP} && \textbf{62.1} & \textbf{86.4} & \textbf{93.0} & \textbf{43.0} & \textbf{71.0} & \textbf{80.3} & \textbf{36.0} & \textbf{61.2} & \textbf{72.3} & \textbf{22.2} & \textbf{45.8} & \textbf{57.3}\\ 
\midrule
CLIP$\Diamond$ &\multirow{2}{*}{ViT-B/16}  & 54.9 & 80.0 & 88.4 & \textbf{37.2} & \textbf{64.3} & 74.3 & 30.7 & \textbf{56.2} & 67.4 & 19.1 & 40.9 & 52.5 \\ 
\textbf{SoftCLIP} & & \textbf{56.2}&\textbf{82.1}&\textbf{88.6}&\textbf{37.2}&\textbf{64.3}&\textbf{74.5}&\textbf{30.9}&\textbf{56.2}&\textbf{68.3}&\textbf{19.2}&\textbf{41.2}&\textbf{52.6}\\ 
\bottomrule
\specialrule{0em}{1pt}{1pt}
\multicolumn{4}{l}{\small $\Diamond$ Our Implementation}\\
\multicolumn{12}{l}{\small $^\dagger$ Tested with: https://github.com/Sense-GVT/DeCLIP\#supported-models}
\end{tabular}}
\end{table*}

Next, we validate the efficacy of our proposed method on image-text retrieval task. To this end, we conduct zero-shot image-text retrieval experiments on the Flikcer30K and MS-COCO datasets, and present the obtained results in Table~\ref{sotaretrieval}. The experimental results demonstrate that SoftCLIP confers significant improvements on both datasets. In particular, when the image encoder is ResNet50, SoftCLIP brings a top-1 hit accuracy improvement of 7.2\% and 5.9\% on Flicker30K image-to-text and text-to-image retrieval tasks respectively. Furthermore, SoftCLIP consistently outperforms DeCLIP pre-trained with the same dataset by a significant margin.

\subsection{Ablation Study}

In this section, we first conduct ablation studies to demonstrate the effectiveness of each module in SoftCLIP, and then explore some other factors which may influence the performance. All the ablation experiments are conducted on CC3M for 8 epochs. More ablation results can be seen in the supplementary materials.

\noindent\textbf{Effectiveness of Each Module}
To verify the effectiveness of each component proposed in SoftCLIP, we conduct a series of experiments with all components added to the CLIP paradigm successively. As demonstrated 
 in Table~\ref{ablationtable}, only the CLIP loss plus the naive soft loss $\mathcal{L}_{\mathrm{soft}}$ can bring significant gains, even exceeding the label smoothing strategy appreciably. Moreover, the adjunction of relation-enhanced soft loss $\mathcal{L}_{\mathrm{soft}}^{\mathrm{re}}$ and the symmetrization of KL-Divergence can further improve the model performance.


\begin{table}[t]
\caption{The effectiveness of each component in SoftCLIP.}
\label{ablationtable}
\small
\setlength{\belowcaptionskip}{1pt}
\centering
\setlength{\tabcolsep}{1mm}{
\begin{tabular}{ccc}
\toprule
\multirow{2}{*}{\textbf{Method}} & \textbf{ResNet50} & \textbf{ViT-B/32}  \\ 
  & \textbf{IN ZS Top-1}  & \textbf{IN ZS Top-1}  \\
\midrule
CLIP (Baseline) & 16.5 & 10.7 \\
 + Label Smoothing & 18.3(\textcolor{darkergreen}{+1.8}) & 11.2(\textcolor{darkergreen}{+0.5}) \\
\hdashline
CLIP + Soft Loss & 20.5(\textcolor{darkergreen}{+4.0}) & 11.7(\textcolor{darkergreen}{+1.0}) \\
+ Relation-enhanced Soft Loss & 21.4(\textcolor{darkergreen}{+4.9}) & 12.2(\textcolor{darkergreen}{+1.5}) \\
+ Symmetric KL (SoftCLIP)  & \textbf{22.1(\textcolor{darkergreen}{+5.6})} & \textbf{12.5(\textcolor{darkergreen}{+1.8})} \\
\bottomrule
\end{tabular}}
\vspace{2pt}
\end{table}

\begin{figure}[t]
    \centering
    \includegraphics[width=8.2cm]{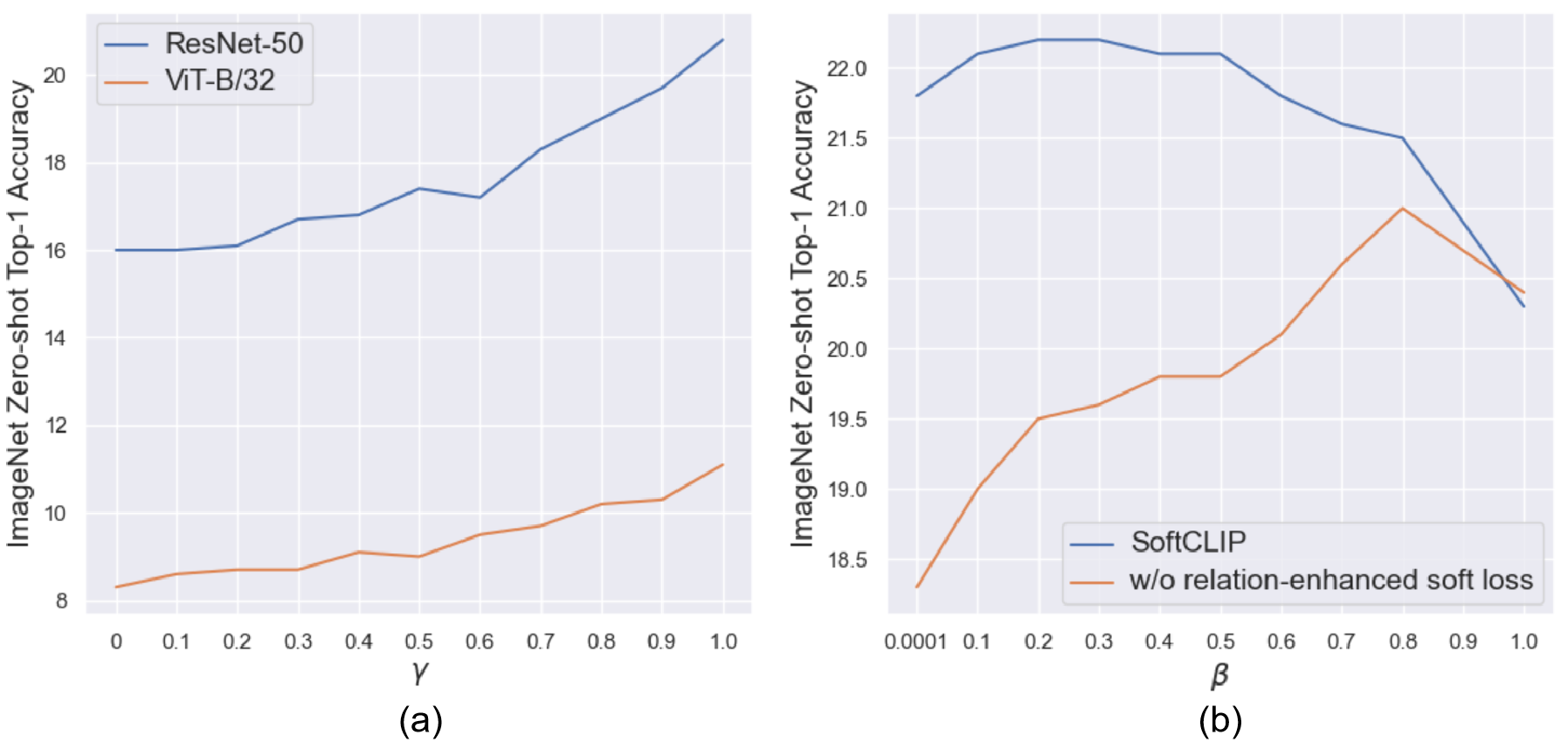}
    \caption{(a) The influence of ROI-tag guidance and image-text guidance at different mixing ratios. (b) The influence of soft self-similarity label and hard one-hot label at different mixing ratios.}
    \label{figure_gvsf}
\end{figure}

\noindent\textbf{Ablation about the Source of Softened Targets}
As we have mentioned in the methodology part, image and text self-similarities are more intuitive to serve as the softened targets compared with ROI and tag self-similarities. Here we provide experimental basis to demonstrate why we choose ROIs and tags. Let $\mathcal{L}(R, A)$ denote the soft loss plus relation-enhanced soft loss under the guidance of ROI and tag self-similarities, and $\mathcal{L}(I, T)$ denote that under the guidance of image and text self-similarities. We additionally experiment with a mixed loss function denoted as
$\mathcal{L}=\gamma\mathcal{L}(R, A) + (1-\gamma) \mathcal{L}(I, T)$, where $\gamma$ is adjustable to control the proportion of the two terms and the CLIP loss is not included in this ablation. The variety of the model performance with respect to $\gamma$ is depicted in Figure \ref{figure_gvsf}(a), which reveals that the model performs better with higher ratio of $\mathcal{L}(R, A)$, \textit{i.e.}, the guidance from ROI and tag self-similarities. We attribute it to two reasons: One is that the image and text similarities are inaccurate in the early training stage, while ROIs and tags inherently contain fine-grained internal alignment due to the priors from the task of object detection; The second reason is that complete images and captions only provide a global understanding, which is relatively coarse, whereas ROIs and tags can capture more detailed local information, providing better guidance.

\begin{figure*}[t]
    \centering
    \includegraphics[width = 0.95\textwidth]{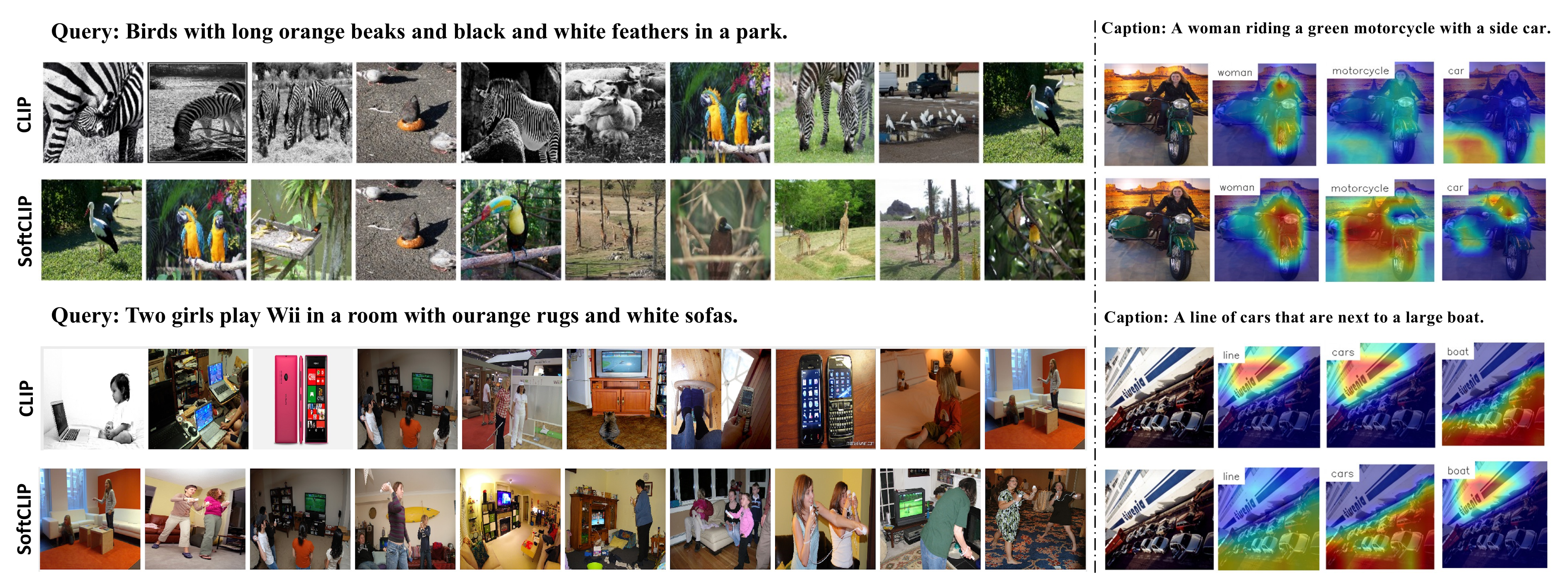}
    \caption{(a) Text-to-image retrieval examples on MS-COCO dataset. From left to right are the top 10 retrieved images from rank1 to rank10. (b) Grad-CAM heatmaps for finding the correspondence from word in the caption to region in the image.}
    \label{cam}
\end{figure*}

\noindent\textbf{Influence of the Parameter $\bm{\beta}$} \; Recall that $\beta$ is the weighting coefficient to mix the one-hot hard label and the soft self-similarity label in Equation (\ref{eqmix1}) (\ref{eqmix2}). Higher value of $\beta$ indicates higher proportion of the self-similarity label. Here we explore the influence of $\beta$, which is shown in Figure \ref{figure_gvsf}(b). In SoftCLIP (see the blue line), the optimal performance is achieved with $\beta$
between 0.1 and 0.5. However, as it increases to $\beta > 0.8$, the performance declines dramatically, which implies that pure self-similarity labels have very poor guidance, hence requiring the reconciliation of hard labels. Another interesting phenomenon is that the accuracy only drops slightly when we mix a very small ratio of the soft label, \textit{i.e.}, $\beta=0.0001$. Our explanation is that the relation-enhanced soft loss term is taking effect. A very small value of $\beta$ (0.0001) leads to a dominant positive logit (more than 0.9999) in the softened target with all the negatives overwhelmed. Nevertheless, the negative logits can be prominent again after being disengaged in the distribution, hence, the model can still capture the relation with negatives. To verify this, we conduct additional experiments with the relation-enhanced soft loss removed (see the orange line). In this configuration, the model performance drops sharply when $\beta<0.2$, which is consistent with the theoretical analysis.


\begin{figure}[t]
    \centering
    \includegraphics[width=8cm]{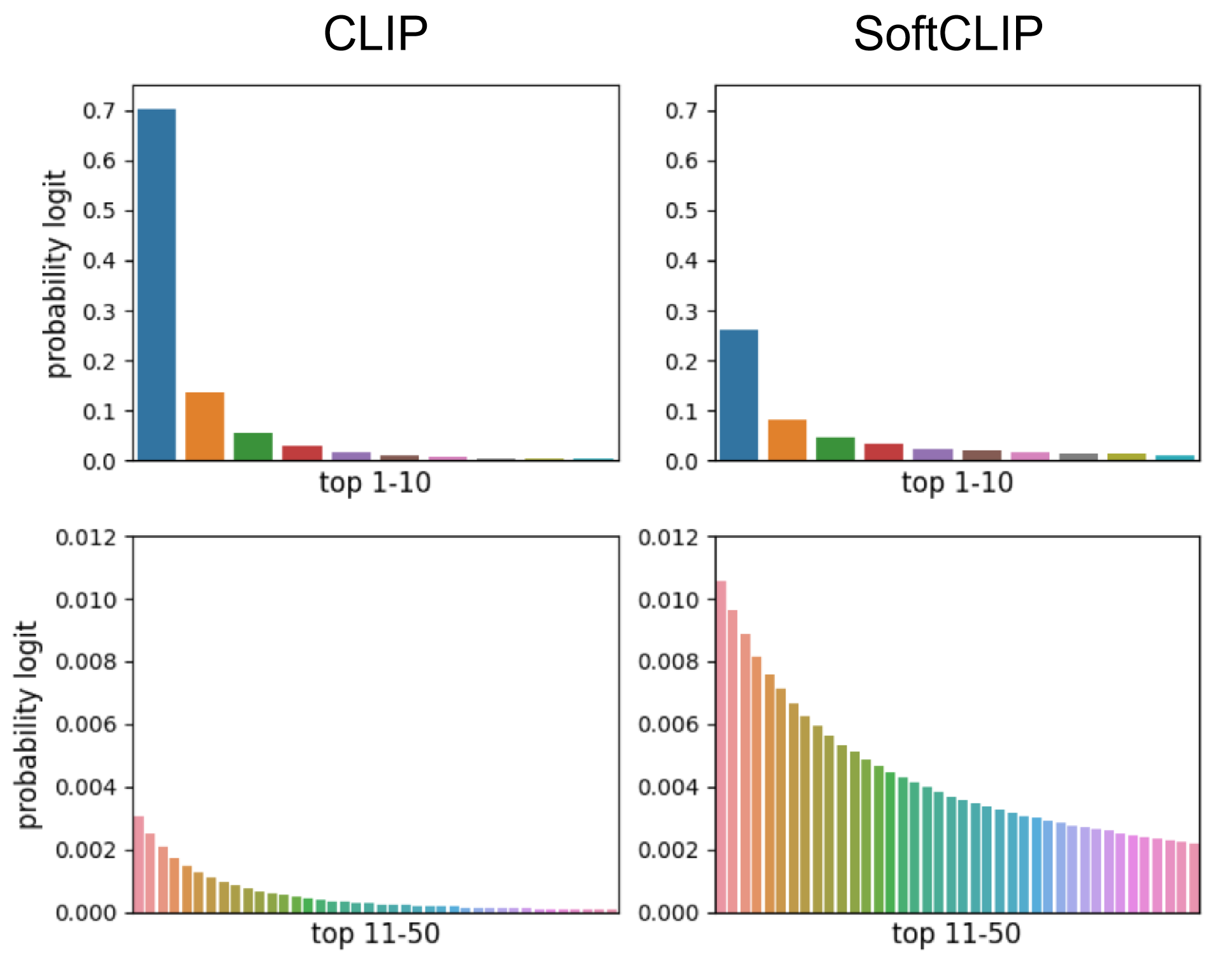}
    \caption{The averaged top 1-50 probability logits obtained through the text-to-image retrieval on Flickr30K, with top 1-10 in the first row and top 11-50 in the second row. In CLIP, the top 1 logit dominates over all others. However, SoftCLIP does not give the top 1 so high a confidence value, but distributes it to other logits for a soft cross-modal relation.}
    \label{figure6}
\end{figure}
\subsection{Visualization}

\noindent\textbf{Text-to-Image Retrieval}~ In Figure~\ref{cam}(a), we give some text-to-image top 10 retrieval results on MS-COCO. It can be seen in the first example that, CLIP tends to narrowly focus on the unitary and specific expression, such as ``\texttt{black and white}", while ignoring others like ``\texttt{birds}", resulting in the retrieval of mostly images of zebras. Whereas, SoftCLIP has a more comprehensive understanding of the text-image relationship and can retrieve the images that have a better match with the query text.

\noindent\textbf{Word-level Localization} Grad-CAM~\cite{selvaraju2017grad} is utilized to show the word-level localization in the image for an image-text pair. As shown in Figure~\ref{cam}(b), SoftCLIP has more precise responses to some nouns compared to CLIP and can accurately locate the region related to the noun. For instance, in the second example, SoftCLIP can exactly locate the corresponding regions of ``\texttt{cars}" and ``\texttt{boat}", while CLIP are confused. We attribute this to the introduction of fine-grained softened target, \textit{i.e.}, the object-level intra-modal self-similarity. 

\subsection{Analysis} 
Here we try to show how SoftCLIP is able to boost soft cross-modal alignment and model many-to-many relationships through text-to-image retrieval on Flickr30K. Specifically, for each text, we obtain its normalized similarity distribution over all the images. The distribution vector are sorted with higher probability logits in front. Across all texts, we can get the average of the sorted vectors consisting of probability values from high to low. Then we pick out top 50 logits and plot them in Figure \ref{figure6} (top 1-10 in the first row and top 11-50 in the second row). It can be found that CLIP is overconfident to the top 1 logit, with top 11-50 overwhelmed. While SoftCLIP sacrifices some confidence in the top 1 and distributes it to the others, allowing for a soft cross-modal relation.

\section{Conclusions}

In this paper, we propose SoftCLIP, a novel approach that relaxes the strict one-to-one constraint and achieves a soft cross-modal alignment by introducing intra-modal self-similarity as softened target and disentangling negatives in the distribution. SoftCLIP can model the commonly existing many-to-many relationships in the web-crawled noisy image-text datasets. Extensive experiments on several tasks  demonstrate the effectiveness of the proposed SoftCLIP.

\appendix
\section{Dataset Details}
\subsection{Pre-training Datasets}
 SoftCLIP is mainly pre-trained on three datasets,  CC3M~\cite{changpinyo2021conceptual}, CC12M~\cite{sharma2018conceptual} and YFCC15M-V2~\cite{li2021supervision}, as listed in Table \ref{train_Data}. CC3M  and CC12M conduct the same
image-text filter pipeline on Internet webpage sources, and the difference is that the filtering method of the latter is more relaxed. The actual size of CC12M is less than the advertised, since not all of the provided URLs are valid. YFCC15M is a commonly used subset of YFCC100M~\cite{thomee2016yfcc100m} and there are mainly two
versions of YFCC15M, V1 and V2. YFCC15M-V1~\cite{thomee2016yfcc100m} is obtained by applying the same filtering rule on
YFCC100M as CLIP~\cite{radford2021learning}, while YFCC15M-V2 is collected by DeCLIP~\cite{li2021supervision} with a different filtering
strategy. In addition to a subset of YFCC100M, the YFCC15M-V2 dataset also contains some additional data crawled
from the Internet and is of higher quality than YFCC15M-V1. 
\begin{table}[htp]
\caption{Pre-training datasets.}
\label{train_Data}
\setlength{\belowcaptionskip}{1pt}
\centering
\setlength{\tabcolsep}{3mm}{
\begin{tabular}{cccc}
\toprule
\textbf{Dataset} & CC3M & CC12M & YFCC15M-V2  \\
\midrule
\textbf{Size} & 3M & 10M & 15M \\
\bottomrule
\end{tabular}}
\end{table}

\subsection{Downstream Zero-shot Classification Datasets}
 We evaluate the transferability of our model on 7 downstream classification datasets,
ImageNet~\cite{deng2009imagenet}, Pets~\cite{data_pets}, Describable Textures~\cite{data_dtd}, Food-101~\cite{data_food}, Flowers-102~\cite{data_flower}, SUN397~\cite{data_sun} and Caltech-101~\cite{data_cal}, as listed in Table \ref{classifcationdatasets}. And we follow the same data split method and evaluation metric as CLIP for fair comparison.


\begin{table}[htp]
\setlength{\belowcaptionskip}{1pt}
\centering
\caption{Datasets for downstream zero-shot classification task.}
\vspace{2pt}
\setlength{\tabcolsep}{1.1mm}{\begin{tabular}{cccc}
\toprule
\textbf{Dataset}& \textbf{Abbreviation}  &  \textbf{Classes}  & \textbf{Test Size}   \\
\midrule
ImageNet & IN & 1000 & 50,000\\
Oxford-IIIT Pets & PETS & 37  & 3,669\\
Describable Textures & DTD & 47  & 1,880 \\
Food-101 &F101 &  101  & 25,250 \\
Oxford Flowers 102 &FLOW & 102  & 6,149 \\
SUN397& SUN & 397 & 19,850 \\
Caltech-101 & CAL& 102 & 6,085 \\
\bottomrule
\end{tabular}}
\label{classifcationdatasets}
\end{table}

\section{More Visualizations}
\begin{figure*}[t]
    \centering
    \includegraphics[width = 1.0\textwidth]{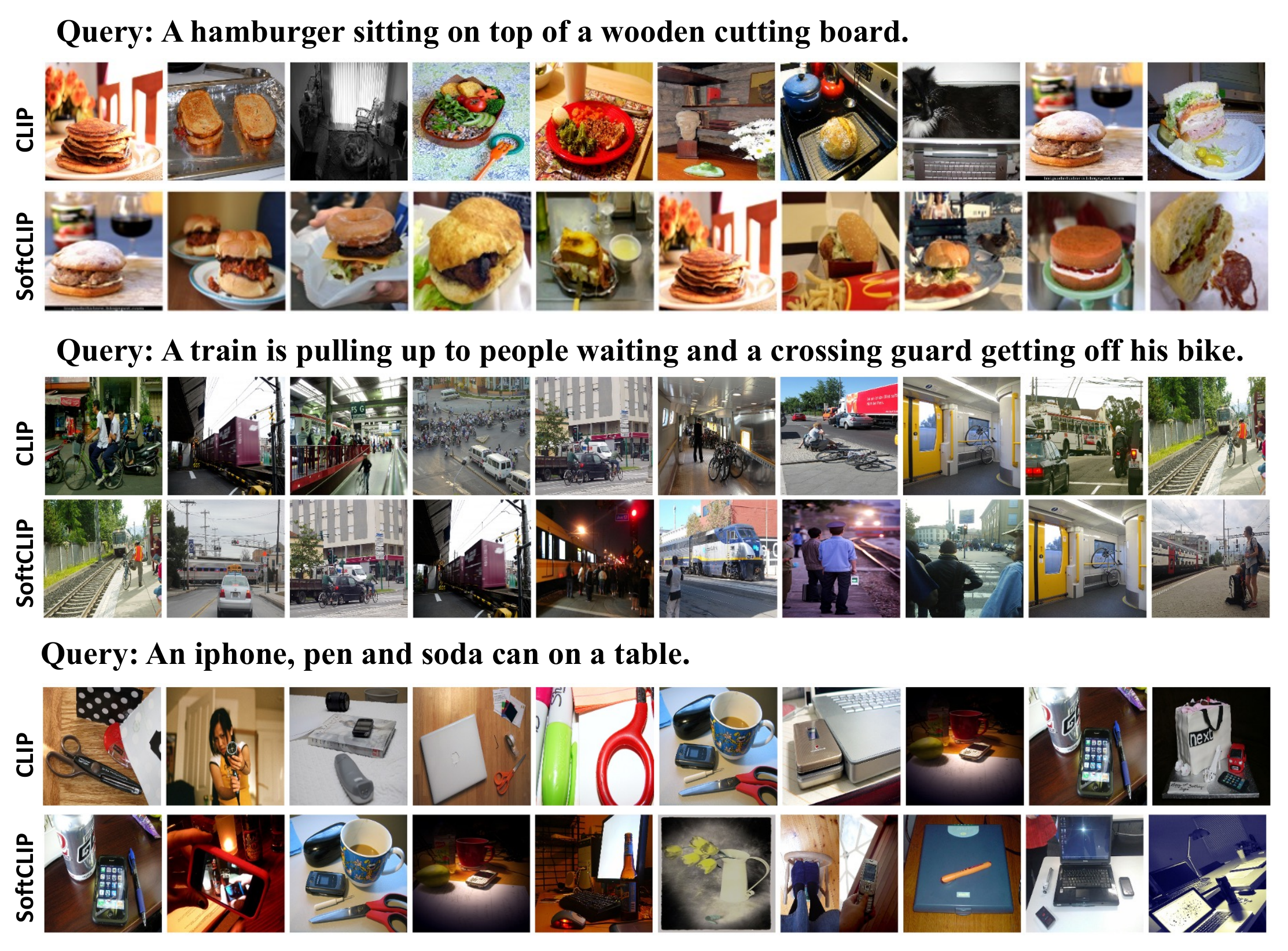}
    \caption{Text-to-image retrieval examples on MS-COCO dataset. From left to right are the top 10 retrieved images from rank1 to rank10.}
    \label{more_vis}
\end{figure*}

Here we provide more text-to-image retrieval examples on MS-COCO~\cite{lin2014microsoft}, as shown in Figure \ref{more_vis}, indicating that SoftCLIP has a more comprehensive understanding of the text-image relationship and can retrieve the images that have a better match with the query text compared with CLIP.

\section{More Ablation}
\subsection{Different Supervision Forms}
In SoftCLIP, we directly use the ROI  self-similarity (R2R) as the soft label the vision-to-language (v2l) loss and tag self-similarity (A2A) in the language-to-vision (l2v) loss. However, there also exist some other supervision forms, as depicted in Table \ref{ablation_sup}. It can be seen that swapping R2R and A2A does not cause performance degradation. Moreover, the performance only drops slightly when replacing R2R and A2A with ROI-to-tag similarity (R2A) and tag-to-ROI similarity (A2R). Therefore, what matters here is the introducing of ROI features and the corresponding tags, rather than the supervision form.
\begin{table}[t]
\caption{ImageNet zero-shot performance with different supervision forms. v2l and l2v denote vision-to-language and language-to-vision respectively. R2R and A2A denote ROI  self-similarity and tag self-similarity respectively. R2A and A2R denote ROI-to-tag similarity and tag-to-ROI similarity respectively.}
\label{ablation_sup}
\setlength{\belowcaptionskip}{1pt}
\centering

\setlength{\tabcolsep}{3mm}{
\begin{tabular}{cccc}
\toprule
\textbf{Method} & \textbf{\tabincell{c}{\textbf{v2l} \\ \textbf{soft label}}} &  \textbf{\tabincell{c}{\textbf{l2v} \\ \textbf{soft label}}}  & \textbf{\tabincell{c}{\textbf{ResNet50} \\ \textbf{IN ZS Top1}}} \\ 
\midrule
 SoftCLIP &  R2R & A2A & \textbf{22.1}       \\
\hdashline
\multirow{3}{*}{\tabincell{c}{Other \\ Forms}} & A2A  & R2R & \textbf{22.1}   \\ 
  & R2A & A2R & 21.9    \\
 &  A2R & R2A  & 21.8  \\ 
\bottomrule
\end{tabular}}
\end{table}

\subsection{Varying the Training Data Size}
\begin{figure}[ht]
    \centering
\includegraphics[width=8.2cm]{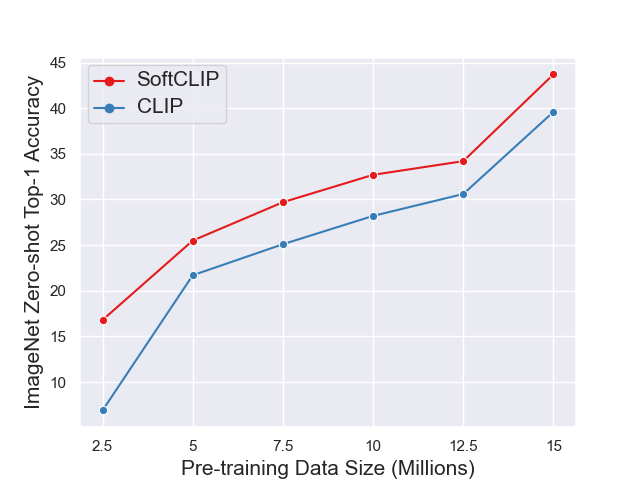}
    \caption{The performance with ResNet50 image encoder on ImageNet zero-shot classification task when varying the training data size of YFCC15M-V2. The subsets are randomly sampled from YFCC15M-V2.}
    \label{figure_data}
\end{figure}

Considering that the noise in pre-training datasets may be alleviated when using sufficient data, we additionally conduct a sweep of experiments with ResNet50 image encoder on YFCC15M-V2, varying the training data size of 2.5, 5, 7.5, 10, 12.5 and 15M. All the subsets are randomly sampled from YFCC15M-V2. As shown in Figure \ref{figure_data}, no matter what the image encoder is, SoftCLIP can maintain a significant improvement over CLIP across all the training data sizes, which indicates the effectiveness and robustness of SoftCLIP.

\subsection{Aggregation of ROI Features}
The ROI feature sequence is feed into the rear part of image encoder in SoftCLIP. That is to say, one or more multi-head self-attention (MHSA)~\cite{dosovitskiy2020image} layers are used to aggregate the ROI features and obtain the final representation vector through an additional \texttt{cls} token. However, considering that ROI features already encode high-level visual semantics, a simpler and more natural way is to get the representation vector through raw ROI features, without MHSA layers processing. Here we experiment with another three aggregation modes. Specifically, for each element in the representation vector, we take the mean/max/min value across all the features in a ROI feature sequence. The results are depicted in Table \ref{ablation_aggre}, which demonstrates that the processed ROI features by MHSA layers in SoftCLIP can bring some gains compared with raw ROI features.
\begin{table}[htp]
\caption{ImageNet zero-shot performance with different aggregation modes of ROI features.}
\label{ablation_aggre}
\vspace{2pt}
\setlength{\belowcaptionskip}{1pt}
\centering
\setlength{\tabcolsep}{3mm}{
\begin{tabular}{ccc}
\toprule
\textbf{\tabincell{c}{\textbf{Aggregation} \\ \textbf{Mode}}} & \textbf{\tabincell{c}{\textbf{ResNet50} \\ \textbf{IN ZS Top1}}} & \textbf{\tabincell{c}{\textbf{ViT-B/32} \\ \textbf{IN ZS Top1}}}  \\
\midrule
 mean & 21.8 & 11.9 \\
max & 21.8 & 12.2\\
min & 19.2 & 10.4 \\
 MHSA (SoftCLIP) & \textbf{22.1} & \textbf{12.5}\\
\bottomrule
\end{tabular}}
\end{table}

{\small
\bibliographystyle{ieee_fullname}
\bibliography{egbib}
}

\end{document}